\documentclass{article}
\usepackage{ijcai20}

\usepackage{times}

\usepackage{soul}
\usepackage{url}
\usepackage[hidelinks]{hyperref}
\usepackage[utf8]{inputenc}
\usepackage[small]{caption}
\usepackage{graphicx}
\usepackage{amsmath}
\usepackage{booktabs}
\usepackage[utf8]{inputenc}
\usepackage{amsmath}
\usepackage{algorithmic}
\usepackage{algorithm}
\usepackage{subcaption}
\usepackage{float}
\usepackage{array}
\usepackage{amsfonts}
\usepackage[hidelinks]{hyperref}
\usepackage[normalem]{ulem}
\useunder{\uline}{\ul}{}
\usepackage{booktabs}
\usepackage{graphicx}
\usepackage{makecell}
\usepackage{units}

\usepackage{graphicx}

\title{Augmented Replay Memory in Reinforcement Learning With Continuous Control}

\author{
Mirza Ramicic$^1$\footnote{Contact Author}\and
Andrea Bonarini$^2$\and
\affiliations
$^1$Artificial Intelligence Center\\
Faculty of Electrical Engineering\\
Czech Technical University in Prague\\
Prague, Czech Republic\\
$^2$AI and Robotics Lab\\ Dipartimento di Elettronica, Informazione e Bioingegneria\\
Politecnico di Milano\\
Milan, Italy\\
\emails
ramicmir@fel.cvut.cz,
andrea.bonarini@polimi.it
}

\begin{document}

\maketitle

\begin{abstract}
Online reinforcement learning agents are currently able to process an increasing amount of data by converting it into a higher order value functions. This expansion of the information collected from the environment increases the agent's state space enabling it to scale up to a more complex problems but also increases the risk of forgetting by learning on redundant or conflicting data. To improve the approximation of a large amount of data, a random mini-batch of the past experiences that are stored in the replay memory buffer is often replayed at each learning step. The proposed work takes inspiration from a biological mechanism which act as a protective layer of human brain higher cognitive functions: active memory consolidation mitigates the effect of forgetting of previous memories by dynamically processing the new ones. The similar dynamics are implemented by a proposed augmented memory replay \textit{AMR} capable of optimizing the replay of the experiences from the agent's memory structure by altering or augmenting their relevance.
Experimental results show that an evolved \textit{AMR} augmentation function capable of increasing the significance of the specific memories is able to further increase the stability and convergence speed of the learning algorithms dealing with the complexity of continuous action domains.
\end{abstract}

\section{ Introduction }
Studies concerning human and animal learning have identified that the process of encoding new memories into long term storage is not so straightforward as previously thought. Recent studies have found that it involves a process of active memory consolidation, or AMC~\cite{diekelmann2010memory,rasch2013sleep,feld2015sleep}, that facilitates a better memory integration into the higher level cortical structures and also prevents forgetting previously encoded information. This process occurs while sleeping, a time when the brain is not encoding or perceiving new stimuli and relies on the memories 
stored in a short-term hippocampal structure when awake. Before their integration in the long term cortical structures, experiences are reactivated or replayed in the hippocampal memory as a part of the active consolidation. The word ``active'' in AMC implies that, in the process of consolidation, memories are altered in a way that their further integration into the existing knowledge wouldn't induce forgetting of the previous ones.

The active structural modification of the consolidated memories is selective and, for the memories that are deemed to be the more important ones it will facilitate strengthening to reach a certain retrieval threshold. However, if the memory trace is deemed not strong enough for some memories it will result in their loss\cite{dumay2016sleep,schreiner2018gain}.

Biological architectures found in human brain and the computational reinforcement learning processes both use a functionally similar mechanism of replay memory. Along the introduction of artificial neural networks, or ANN, as function approximators in temporal-difference, or TD, learning~\cite{lin1993reinforcement}, the techniques that aim at their efficient training most commonly use a replay buffer of previous experiences out of which a mini-batch is sampled for re-learning at each time step. This technique has been recently revived in Deep Q-learning\cite{mnih2013playing,mnih2015human}. Since in TD approaches the ANN is constantly updated to better represent the state-action value pairs $Q(s,a)$, which govern the agent's policy $\pi$, the mechanisms involved in its training such as mini-batch replay became increasingly influential to the learning process itself.

Another advantage of the replay memory structure is that, when implemented, it acts as a form of agent's cognition: depending on the way it is populated, it can alter how the agent perceives the information. In this way, a learning agent is not only concerned about the information it receives from its immediate environment, but also about the way in which this information is interpreted by this cognitive mechanism.

In the proposed approach an effective, but simple, mechanism of replay memory is extended with the ability to actively and dynamically process the information during the replay and thus bringing it closer to the functional characteristics of actual biological mechanisms. The dynamic processing mechanism of \textit{Augmented Memory Replay} or \textit{AMR} presented here is inspired by human \textit{active memory consolidation} and it is capable of altering the importance of specific memories by altering their reward values, thus mimicking the AMC's process of deeming the memory above the retrieval threshold. In the experiments reported in this paper, the \textit{augmentation} dynamics are evolved over generations of learning agents performing reinforcement learning tasks in various environments. Their fitness function is defined in a straightforward way as their cumulative performance over a specific environment.
Experimental results indicate that \textit{AMR} type of memory buffer shows an improvement in learning performance over the standard static replay method in all of the tested environments. 

\section{Related Work}
An extension of DDPG algorithm was proposed by Hausknecht and Stone~\cite{hausknecht2015deep} allowing it to deal with a low level parameterized-continouos action space. However the evaluation of the approach was limited to a single simulated environment of RoboCup 2D Half-Field-Offense~\cite{hausknecht2016half}.
Hoothoft et. al~\cite{houthooft2018evolved} proposed a meta-learning approach capable of evolving a specialized loss function for a specific task distribution that would provide higher rewards during its minimization by stochastic gradient descent. The algorithm is capable to produce a significant improvement of the agent's convergence to the optimal policy but as its the case with the \textit{AMR} approach the evolved improvements are task specific.

In contrast with the distributed methods like Apex which was proposed by Horgan et. al~\cite{horgan2018distributed} which rely on a hundreds of actors learning in their own instance of the environment \textit{AMR} algorithm works with a single instance just like vanilla DDPG~\cite{lillicrap2015continuous} does. This fact has a significant impact on the computational time a specific algorithm induces to the problem.

Wang et. al~\cite{wang2016sample} introduced an approach that is combining the importance or prioritized sampling techniques together with stochastic dueling networks in order to improve the convergence of some continuous action tasks such as Cheetah, Walker and Humanoid.

Another improvement of a vanilla DDPG is presented by Dai et. al~\cite{dai2017boosting} as Dual-Critic architecture where the critic is not updated using the standard temporal-difference algorithms but it's optimized according to the gradient of the actor.

An approach by Pacella et. al~\cite{pacella2017basic} evolved basic emotions such as fear, used as a kind of motivational drive that governed the agent's behavior by directly influencing action selection. Similar to the \textit{AMR} approach a population of virtual agents were tested at each generation. In this process, each of the agents evolved a specific neural network that was capable of selecting its actions based on the input; this consisted of temporal information, visual perception and good and bad sensation neurons. Over time, the selection of best performing agents gave rise to populations that adopted specific behavioral drives such as being cautious or fearful as a part of a survival strategy. Contrary to the \textit{AMR} which evolves a cognitive mechanism which only complements the main learning process, in~\cite{pacella2017basic} the genetic algorithm represents the learning process itself.

Another evolutionary approach that is used to complement the main reinforcement learning algorithm was presented in~\cite{singh2010intrinsically}. Similarly to \textit{AMR}, it uses a genetic algorithm to evolve an \textit{optimal reward function} which builds upon the basic reward function in a way that maximizes the agent's fitness over a distribution of environments. Experimental results show the emergence of an intrinsic reward function that supports the actions that are not in line with the primary goal of the agent.
\cite{schembri2007evolution} also presented an a reinforcement learning approach which relied on a evolved reinforcer in order to support learning atomic meta-skills. The reinforcement was evolved in a \textit{childhood} phase, which equipped the agents with the meta-actions or skills for the use in the \textit{adulthood} phase.

Persiani et al.~\cite{persiani2018working} proposes a cognitive improvement through the use of replay memory structure like \textit{AMR}. The algorithm makes it possible to learn which chunks of agent's experiences are most appropriate for replay based on their ability to maximize the future expected reward.

A cognitive filter structure was proposed by Ramicic and Bonarini~\cite{ramicic2019selective} able to improve the convergence of  temporal-difference learning implementing discrete control rather than a continuous one. It was able to evolve the ANN capable to select whether a specific experience will be sampled into replay memory or not. Unlike \textit{AMR} this approach did not modify the properties of the experiences.

\section{Theoretical Background}
\subsection{Temporal-difference learning}
The goal of a reinforcement learning agent is to constantly update the function which maps its state to their actions i.e. its policy $\pi$ as close as possible to the \textit{optimal policy} $\pi^*$. The \textit{optimal policy} is a policy that selects the actions which maximize the future expected reward of an agent in the long run \cite{sutton1998reinforcement} and it is represented by a function, possibly approximated by an \textit{Artificial Neural Network} or ANN. The process of updating the policy is performed iteratively after each of the consecutive discrete time-steps in which the agent interacts with its environment by executing its action $a_t$ and gets the immediate reward scalar $r_t$ defined by the \textit{reinforcement function}. This iterative step is defined as a transition over \textit{Markov Decision Process}, and represented it by a tuple $ \left ( s_t,a_t,r_t,s_{t+1} \right ) $. After each transition the agent corrects its existing policy $\pi$ according to the optimal action-value function shown in \autoref{eq:optimal} in order to maximize its expected reward within the existing policy. In the approaches that deal with discrete action spaces, such as \cite{watkins1992q}, the agent can follow the optimal policy $\pi^*$ by taking an optimal action $a^*(s)$ which maximizes the optimal action-value function $Q^*(s,a)$ defined by \autoref{eq:action}.

\begin{equation} Q^*(s,a) = \max_{\pi}\mathbb{E}[R_t \vert s_t = s,a_t = a,\pi]\label{eq:optimal}\end{equation}

\begin{equation}
    \mu(s) = a^*(s) = \max_{a}Q^*(s,a)
    \label{eq:action}
\end{equation}

\begin{equation} Q^\pi(s,a) = \mathbb{E} \left [ r + \gamma 
\max_{a'}Q^\pi(s',a') \vert s,a \right ] 
\label{eq:bellman}
\end{equation}

The correction update to the policy $\pi$ starts by determining how wrong the current policy is with respect to the expectation, or value for the current state-action pair $Q(s,a)$. In case of a discrete action space this expectation of return is defined by the Bellman-optimality equation \autoref{eq:bellman} and it is basically the sum of the immediate reward $r$ and the discounted prediction of a maximum Q-value, given the state $s`$ over all of the possible actions $a`$.
\subsection{Going continuous}
Maximizing over actions in \autoref{eq:action} is not a problem when facing discrete action spaces, because the Q-values for each of the possible actions can be estimated and compared. However, when coping with continuous action values this approach is not realistic: we cannot just explore brute force the values of the whole action space in order to find the maximum.
The more recent approach of~\cite{lillicrap2015continuous} eliminates the maximization problem by approximating the optimal action $a^*(s)$ and thus creating a deterministic policy $\mu(s)$ in addition to the optimal state-value function $Q^*(s,a)$.  Taking the new approximated policy into consideration the Bellman-optimality equation takes the form of \autoref{eq:deterministic} and avoids the inner expectation.

\begin{equation} Q^\mu(s,a) = \mathbb{E} \left [ r + \gamma 
Q^\mu(s',\mu(s')) \vert s,a \right ] 
\label{eq:deterministic}
\end{equation}

In common among all the before mentioned approaches is the concept of \textit{temporal difference}, or TD error, which is basically a difference between the current approximate prediction and the expectation of the Q value. The learning process performs an iterative reduction of a TD error using Bellman-optimality equation as a target, which guarantees the convergence of the agent's policy to the optimal one given an infinite amount of steps \cite{sutton1998reinforcement}.

\subsection{Function approximation}
In order to deal with the increasing dimensionality and continuous nature of state and action spaces imposed by the real-life applications the aforementioned algorithms depend heavily on approximate methods usually implemented using ANN. A primary function approximation makes it possible to predict a $ Q $ value for each of the possible actions available to the agent by providing an agent's current state as input of the ANN. After each time step, the expected Q value is computed using \autoref{eq:deterministic}, and then compared to the estimate that the function approximator provides as its output $ Q(s,a;\Theta) \approx Q^*(s,a) $ by forwarding the state s0 as its input. The difference between the previous estimate of the approximator and the expectation is the TD error. This discrepancy is actually a loss function $ L_i(\Theta_i) $ that can bi minimized by performing a \textit{stochastic gradient descent} on the parameters $\Theta$ in order to update the current approximation of $ Q^*(s,a) $ according to \autoref{eq:gradient}:

\begin{equation} \label{eq:gradient} \nabla_{\Theta_i}L_i(\Theta_i) = 
\left ( y_i - Q(s,a;\Theta_i) \right )\nabla_{\Theta_i}Q(s,a;\Theta_i),
\end{equation}
where \begin{math} y_i = r + \gamma 
Q^\mu(s',\mu(s'));\Theta_{i-1}) \end{math} is in fact the Bellman equation defining the target value which depends on an yet another ANN that approximates the policy function $\mu(s)$ in policy-gradient approaches such as~\cite{lillicrap2015continuous}.
The update to the policy function approximator $\mu_{\Theta}(s)$ is more straightforward as it is possible to perform a gradient ascent on the respective network parameters $\Theta$ in order to maximize the $Q^\mu(s,a)$ as shown in \autoref{eq:policy}.

\begin{equation}
    \max_{\theta} \underset{s \sim {\mathcal D}}{{\mathrm E}}\left[ Q^\mu(s, \mu_{\theta}(s)) \right]
    \label{eq:policy}
\end{equation}

\section{Model Architecture and Learning Algorithm}
In this section we propose a new model that combines the learning approaches of genetic algorithm with reinforcement learning order to improve the convergence of the latter.
For clarity, the proposed model is separated in two main functional parts: evaluation and evolution.

The evaluation part is defined as a temporal-difference reinforcement problem where a reward function is dynamically modified by the proposed \textit{AMR} block. \textit{AMR} is a function approximator implemented by an ANN, which receives in input characteristics of experience that is perceived by the learning agent, and outputs a single scalar value , used to modify the reinforcement value of the transition.

The architecture of the AFB neural network approximator $ (f) $ consists of three layers: three input nodes fully connected to a hidden layer of four nodes, in turn connected to two softmax nodes to produce the final classification. This ANN is able to approximate the four parameters of the experience, respectively given in input, as TD error, reinforcement $ r_i $, entropy of the starting state $ s_t $, and entropy of the next transitioning state $ s_{t+1} $, to a regression output layer that provides a scalar \textit{augmentation} value or $ A_t(s_t,s_{t+1},r_t) $.
The \textit{augmentation} process alters the reward value of each transition by an \textit{augmentation rate} or $\beta$ as shown in \autoref{eq:update}.

\begin{equation}
r_t := r_t + \beta A_t
\label{eq:update}
\end{equation}

\begin{figure*}[h]
\centering
   \includegraphics[width=1\textwidth]{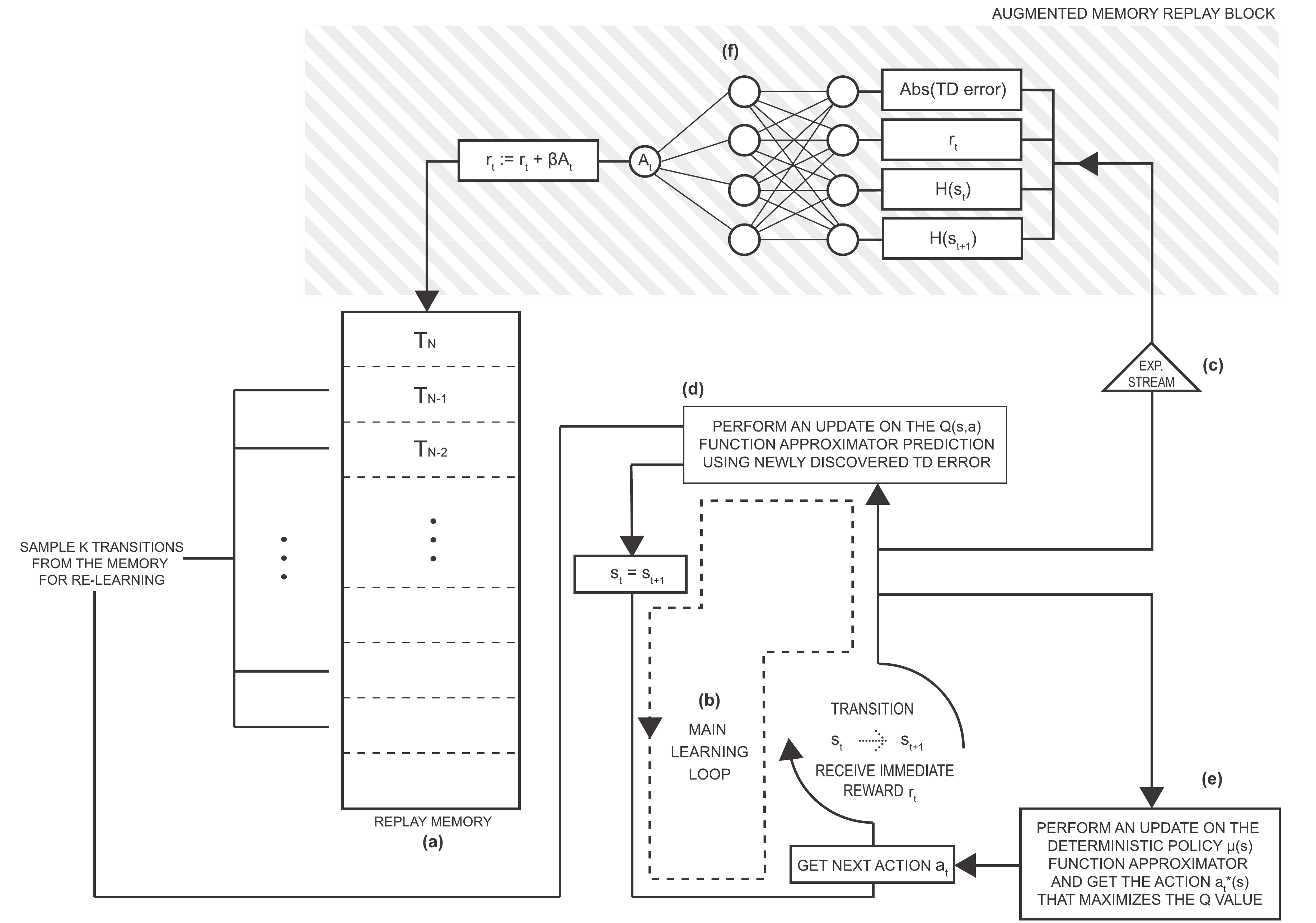}
   \caption{General learning model architecture including the \textit{attention focus block}: $ (a) $ Replay memory implemented as a sliding window buffer of N experiences; $ (b) $ Main learning loop which consists of: $ 1) $ the transition in which the agent performs an action, receiving an immediate reward $ r_t $, while transitioning to a next state $ s_{t+1} $; $ 2) $ performing an update on main function approximator ANN $ (d) $ by back propagating the TD error as a gradient of the $ a_t $ output; $ 3) $ shifting the states for the next iteration in which the $ s_t $ becomes our $ s_{t+1} $; $ 4) $ Updating deterministic policy $ \mu(s) $ by changing the weights $\Theta^\mu$ of the actor ANN approximator $(e)$; $ 5) $ Getting the next action $a_{t+1}$ which maximizes the $Q$ value for the given state $s_t$ according to the actor ANN  $ (d) $ A block implementing Q-value function approximator taking the starting state $ s_t $ on the input and predicting Q-values for each of the available actions on its output; $ (c) $ Raw stream of the experiences that are perceived representing unfiltered cognition of an agent; $ (f) $ Augmented Memory Replay block consisting of an ANN that approximates the augmentation parameter $A_i$ based on the properties of the experience.}
   \label{fig:model}
\end{figure*}


\begin{figure}[h]
\centering
   \includegraphics[width=0.5\textwidth]{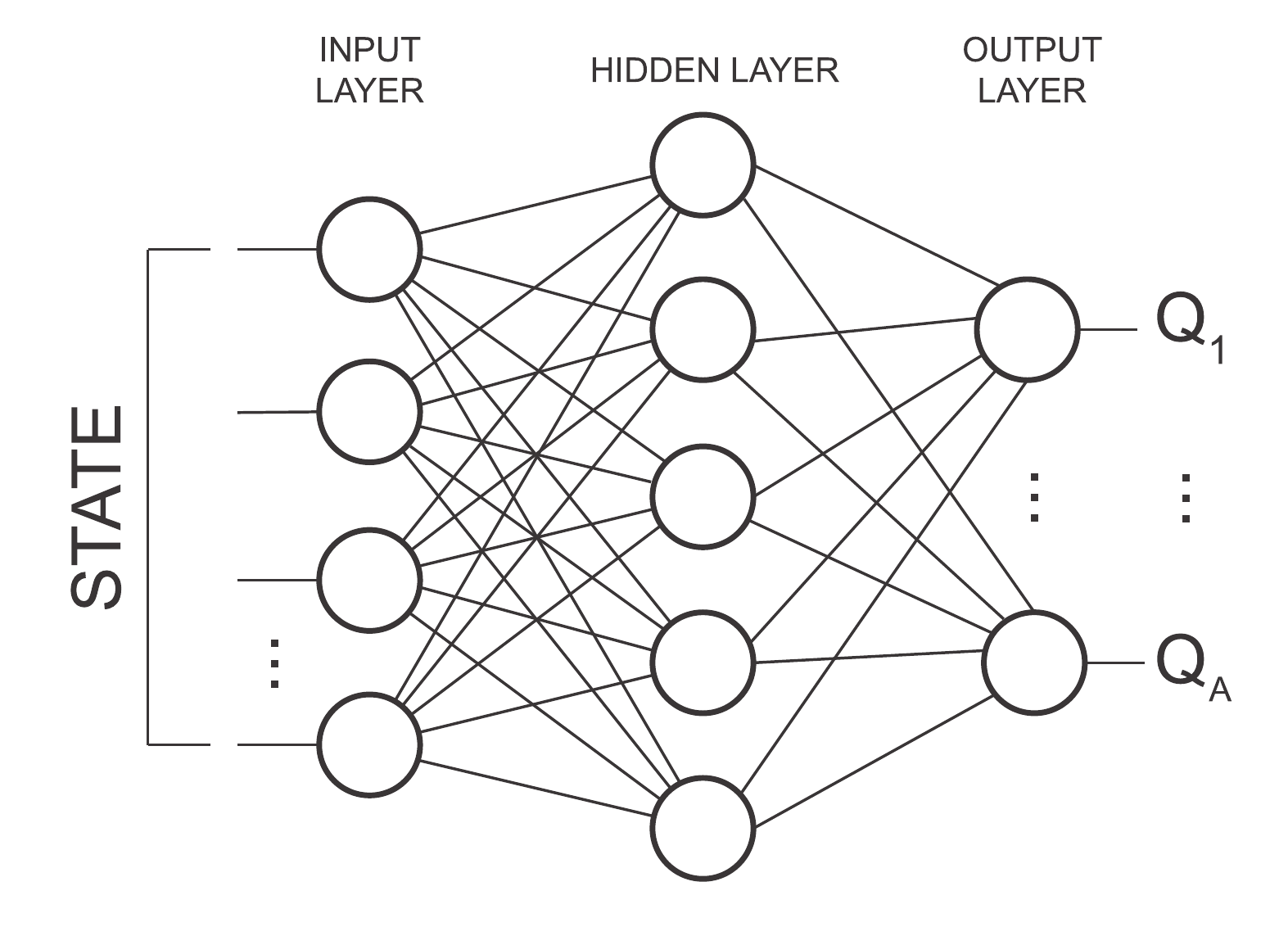}
   \caption{Main function approximator ANN implemented in the (d) block of \autoref{fig:model}: it receives an N-dimensional state as its input and approximates it to $ Q $ values of each of $ A $ possible actions available to an agent at its output, therefore providing an approximation for $ Q(s,a) $ pairs. }
   \label{fig:neural}
\end{figure}

While altering the reward scalar $r_t$ the \textit{AMR} block is able to precisely and dynamically change the amount of influence each transition exerts on the learning process ant thus mimic the aforementioned biological processes~\cite{diekelmann2010memory,rasch2013sleep,feld2015sleep}.

The second component of the proposed architecture evolves the \textit{AMR} block using a \textit{genetic algorithm}, or $ GA $, in order to maximize its fitness function which is represented by the total learning score received by an agent during its evaluation phase. 



\begin{algorithm}[h]                      
\caption{DDPG with Augmented Memory}
\label{a:alg1}
\begin{algorithmic}  
\STATE Initialize critic network $Q(s,a\vert\Theta^Q)$, actor network $\mu(s\vert\Theta^\mu)$ and augmentation network $A(s,r\vert\Theta^\beta)$ with random weights $\Theta^Q$, $\Theta^\mu$ and $\Theta^\beta$
\STATE Initialize target network $Q'$ and $\mu'$ with weights $\Theta^{Q'} \leftarrow \Theta^Q$ and $\Theta^{\mu'} \leftarrow \Theta^\mu$
\STATE Initialize replay buffer $R$
\FOR{episode = 1, M}
\STATE Initialize a random process $N$ for action exploration
\STATE Observe initial state $s_1$
\FOR{t=1, T}
\STATE Select action $a_t=\mu(s_t\vert\Theta^\mu) + N_t$ according to the current policy and exploration noise
\STATE Execute action $a_t$ and observe reward $r_t$ and new state $s_{t+1}$
\STATE Augment the reward $r_t\leftarrow r_t + A_t(s_t,s_{t+1},r_t)$ according to the augmentation network parameters $\Theta^\beta$
\STATE Store transition $(s_t,a_t,r_t,s_{t+1})$ in R
\STATE Sample a random minibatch of $S$ transitions $(s_t,a_t,r_t,s_{t+1})$ from R
\STATE Set $y_i = r_i + \gamma Q'(s_{i+1},\mu'(s_{i+1}\vert\Theta^{\mu'})\vert\Theta^{Q'})$
\STATE Update critic by minimizing the loss $L=\frac{1}{S}\sum_{i}(y_i - Q(s_i,a_i\vert\Theta^{Q'}))^2$
\STATE Update the actor policy using the sampled policy gradient
\STATE $\nabla_{\Theta^{\mu}}J\approx\frac{1}{S}\sum_{i}\nabla_{a}Q(s,a\vert\Theta^Q)\vert_{s=s_i,a=\mu(s_i)}\nabla_{\Theta^{\mu}}\mu(s\vert\Theta^\mu)\vert_{s_i}$
\STATE Update the networks
\STATE $\Theta^{Q'}\leftarrow\tau\Theta^Q+(1-\tau)\Theta^{Q'}$
\STATE $\Theta^{\mu'}\leftarrow\tau\Theta^\mu+(1-\tau)\Theta^{\mu'}$

\ENDFOR
\ENDFOR
\end{algorithmic}
\end{algorithm}

\linespread{1.0}

\section{Experimental Setup}
\subsection{Environment}
\begin{figure*}[h!] 
 \label{fig:env}
  \begin{minipage}[b]{0.25\linewidth}
    \centering
    \includegraphics[width=.9\linewidth]{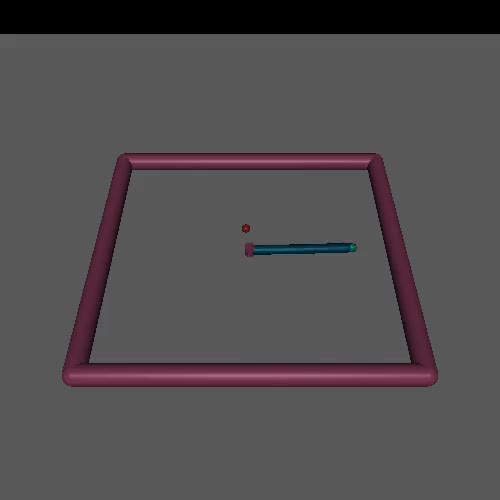} 
    \caption{Reacher} 
    \vspace{4ex}
  \end{minipage}
  \begin{minipage}[b]{0.25\linewidth}
    \centering
    \includegraphics[width=.9\linewidth]{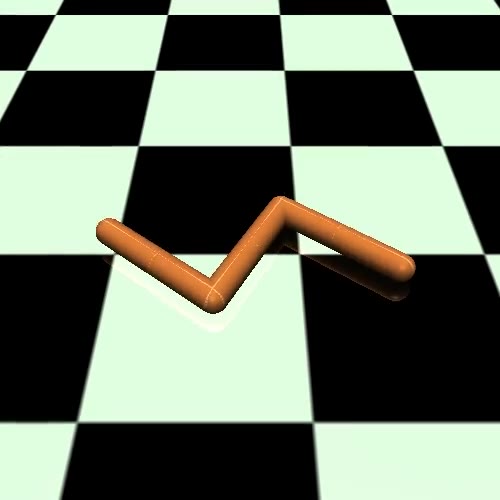} 
    \caption{Swimmer} 
    \vspace{4ex}
  \end{minipage}
  \begin{minipage}[b]{0.25\linewidth}
    \centering
    \includegraphics[width=.9\linewidth]{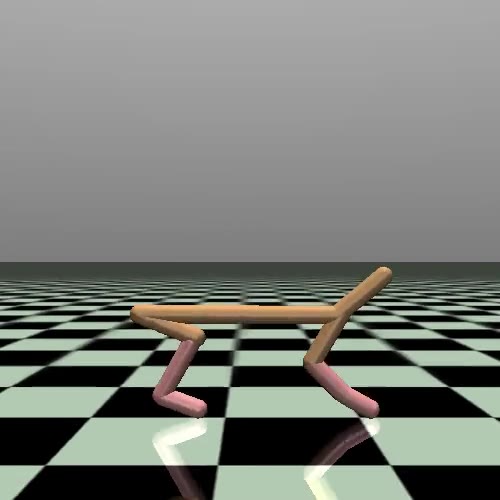} 
    \caption{Half Cheetah} 
    \vspace{4ex}
  \end{minipage}
  \begin{minipage}[b]{0.25\linewidth}
    \centering
    \includegraphics[width=.9\linewidth]{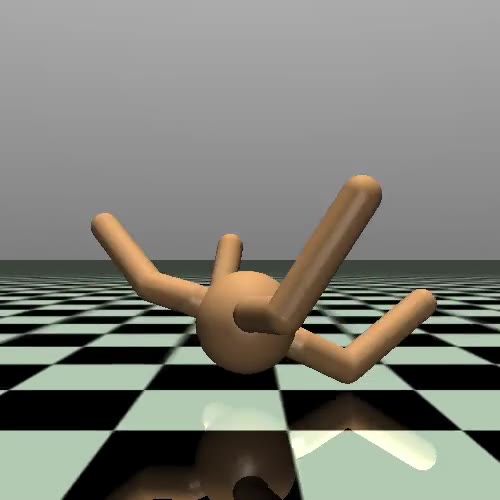} 
    \caption{Ant} 
    \vspace{4ex}
  \end{minipage}
  \caption{Variety of OpenAI Gym environments considered in evaluation. Ordered from low to high complexity.}
\end{figure*}

The evaluation phase applied the proposed variations of the $DDPG$ learning algorithm to a variety of continuous control tasks running on an efficient and realistic physics simulator as a part of OpenAI Gym framework~\cite{1606.01540} and shown in \autoref{fig:env}. The considered environments range from a relatively simple 2D robot (\textit{Reacher-v2}), with a humble 11-dimensional state space, to a complex four-legged 3D robot such as \textit{Ant-v2}\cite{schulman2015high}, which boasts a total of 111-dimensional states coupled with 8 possible continuous actions. Various different tasks of intermediate complexity like making a 2D animal robot run (\textit{HalfCheetah-v2}), and making a 2D snake-like robot move on a flat surface (\textit{Swimmer-v2}~\cite{coulom2002reinforcement}) have also be faced.

\subsection{Function Approximation}
An approximation of $ Q(s,a;\Theta) \approx Q^*(s,a) $ has been implemented using an ANN with one hidden fully connected layer of 50 neurons, able to take an agent's state as an input and produce as output the Q values of all the actions available to the agent. The learning rate of an \textit{critic} $Q$ approximator $\alpha$ is set to $0.002$.

The \textit{actor} function approximator of $ a(s;\Theta) \approx a^*(s) $ is implemented using one hidden dense layer of 30 neurons which outputs a deterministic action policy based on the agent's current state. The \textit{actor} ANN has been trained using slightly higher learning rate of $0.001$ compared to the critic one.

The architecture of the \textit{AMR} function approximator consists of three layers: four input nodes connected to a fully connected hidden layer of four nodes, in turn connected to a single regression node able to produce an \textit{augmentation scalar} as output. This ANN is able to approximate four parameters of the current agent's experience, respectively given in input as an absolute value of TD error, reinforcement, entropy of the starting state $ s_t $, entropy of the transitioning state $ s_{t+1} $, to scalar value $A_t$ that indicates how important the specific experience it to the learning algorithm. 

\subsection{Meta Learning Parameters}
During the evaluations phase at each learning step a batch of 32 experiences were replayed from the fixed capacity memory buffer of 10000. Learning steps per episode were limited to a maximum of 2000. Reward discount factor $\gamma$ was set to a high $0.9$ and soft replacement parameter $\tau$ was $0.01$.
In order to achieve action space exploration an artificially generated noise is added to the deterministic action policy which is approximated by the \textit{actor} ANN. The noise is gradually decreased or adjusted linearly from an initial scalar value $3.0$ to $0.0$ towards the end of the learning.

\section{Experimental Results}
\begin{figure}[h]
\centering
   \includegraphics[width=0.5\textwidth]{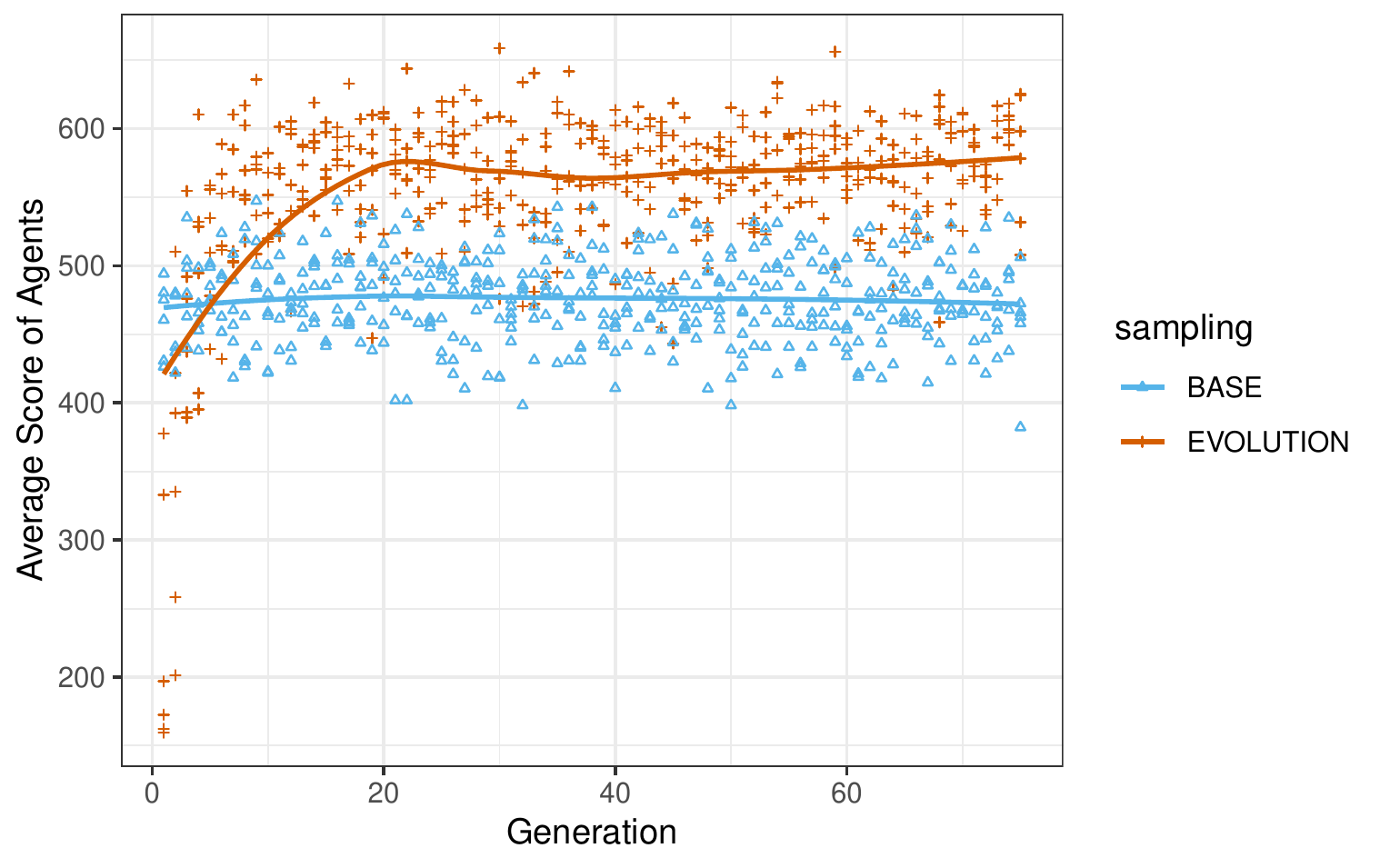}
   \caption{Average score or total reinforcement in Ant environment received over 75 generations of learning agents.}
   \label{fig:ant}
\end{figure}

\begin{figure}[h]
\centering
   \includegraphics[width=0.5\textwidth]{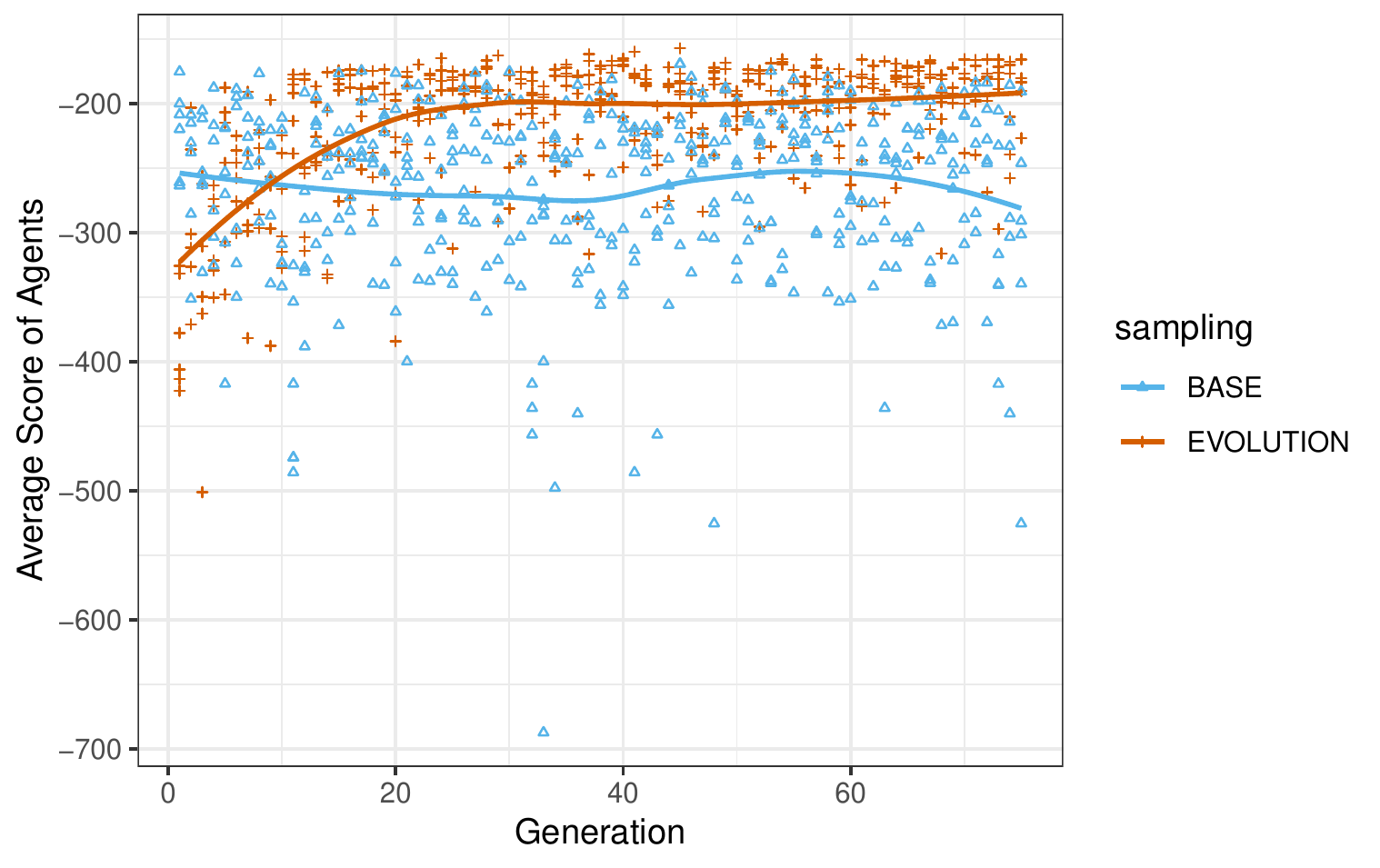}
   \caption{Average score or total reinforcement in Reacher environment received over 75 generations of learning agents.}
   \label{fig:reacher}
\end{figure}

\begin{figure}[h]
\centering
   \includegraphics[width=0.5\textwidth]{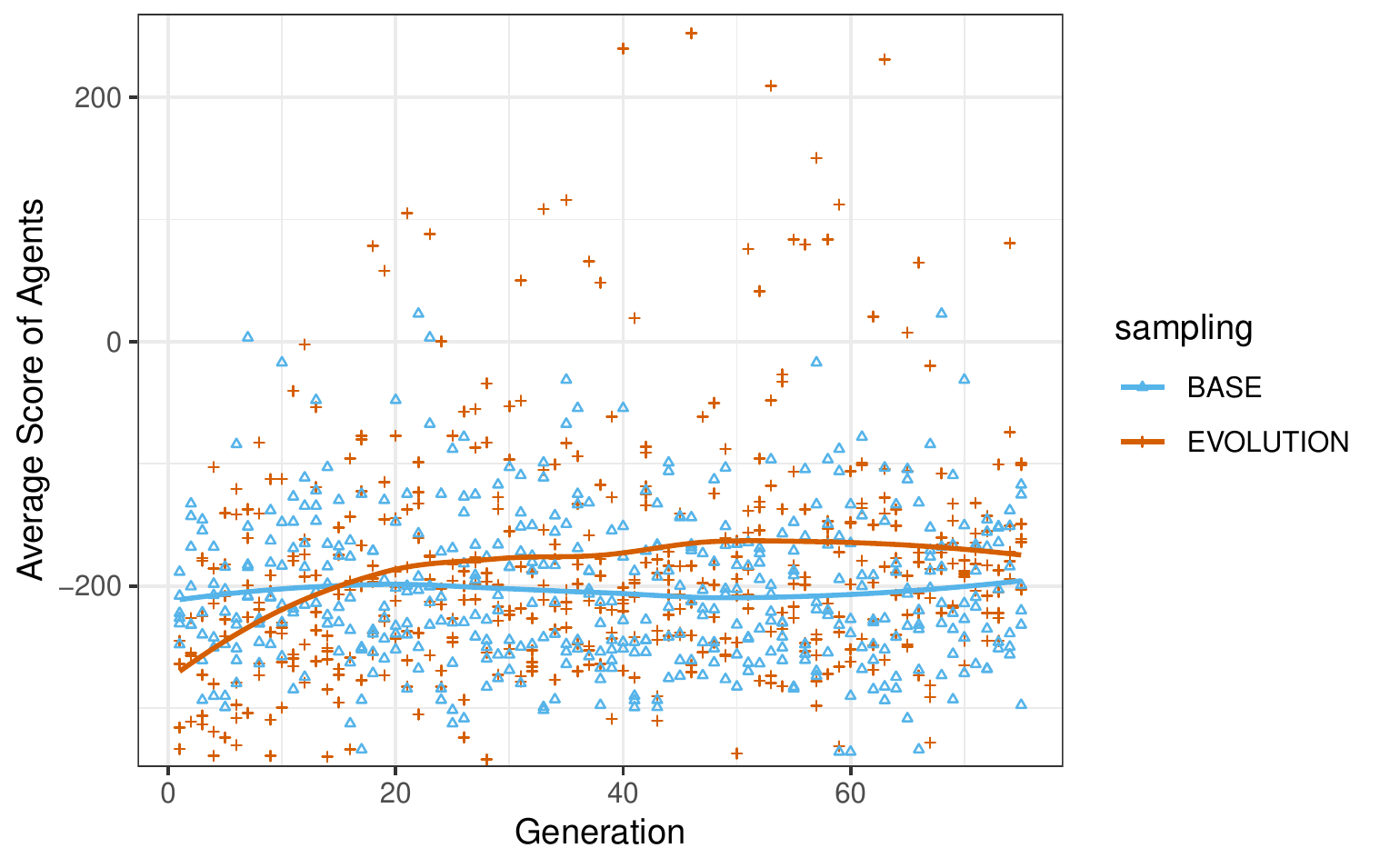}
   \caption{Average score or total reinforcement in Cheetah environment received over 75 generations of learning agents.}
   \label{fig:cheetah}
\end{figure}

\begin{figure}[h]
\centering
   \includegraphics[width=0.5\textwidth]{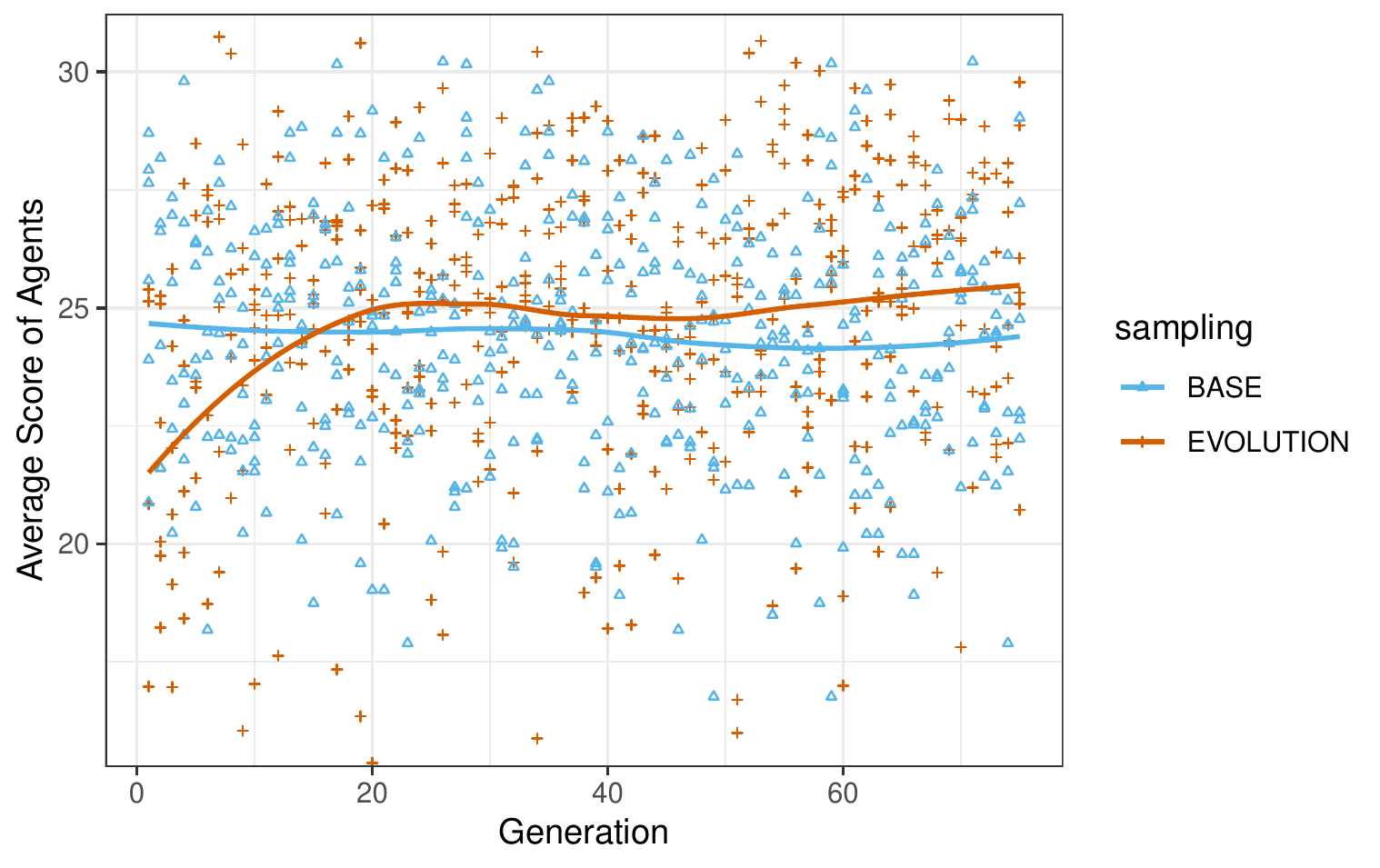}
   \caption{Average score or total reinforcement in Swimmer environment received over 75 generations of learning agents.}
   \label{fig:swimmer}
\end{figure}

The proposed algorithm evolved the \textit{AMR's} neural network weights $\Theta^{AMR}$ trough a total of 75 generations. At each generation, the learning performance of 10 agents were evaluated based on their their total cumulative score during 200 learning episodes. Only the best 5 scoring agents of each generation had an opportunity to propagate their genotypes to the next generation in order to form a new population. As shown in \autoref{fig:ga} this process involved common $GA$ techniques such as crossover and random mutation. The crossover of the genotypes, which are actually the \textit{AMR} weights, were prioritized based on the agents cumulative score and the mutation was additionally applied at a rate of $0.25$ by adding a random scalar between $0.1$ and $-0.1$ to the weights.
The obtained experimental results which are presented along the Figures~\ref{fig:ant},~\ref{fig:reacher},~\ref{fig:cheetah},~\ref{fig:swimmer} indicate that the most complex setup of \textit{Ant-v2} improved its learning performance the most when using the proposed \textit{AMR} approach when compared to the baseline approach that have not used \textit{memory augmentation}. Regardless of the environment, it i evident that the proposed evolutionary approach with memory augmentation underperforms at the very first generations but quickly surpasses the baseline in less than 10 generations and further improves the total score of the agent in the following generations. As we can see from~\ref{fig:ant} the \textit{AMR} evolutionary approach improves the \textit{Ant's} quad-legged robot learning about how to walk by a total of $18.9\%$ towards the end of the 75th generation.
\textit{AMR} algorithm have also showed a significant improvement in \textit{Reacher}: the simplest of the environments. In this task the proposed evolutionary approach made the 2D robot hand with one actuating joint learn to fetch a randomly instantiated target faster and produced a $35.4\%$ increase in agent's total cumulative score.
Although not as significant as in \textit{Ant} and \textit{Reacher} setups, the \textit{AMR} approach is also able to improve the performance in the \textit{Cheetah} and \textit{Swimmer} environments as evident from the \autoref{fig:cheetah} and \autoref{fig:swimmer}, respectively.
We can also notice the difference of the score variance between the setups which can be attributed to distinctive robot/environment characteristics; while \textit{Ant} and \textit{Reacher} show relatively low variance in their scores, other problems like \textit{Cheetah} and \textit{Swimmer} have a very high variance.

\section{Discussion}
The presented approach represents yet another inspiration from biological systems, which implements a biologically inspired mechanisms that enables artificial learning agents to better adapt to a specific environment by selectively increasing the relevance of the information perceived. An agent implementing an \textit{AMR} neural network is able to evolve its memory augmentation criteria to best fit the environment that is facing, in few generations. The evolved \textit{AMR's} augmentation criteria modifies the relevance of the information that an agents collects from its immediate environment into its replay memory allowing it to use the same data in a more efficient way during the learning process; this yields a direct improvement in the performance.

Thus, augmenting memory allows for the emergence of an \textit{artificial cognition} as a intermediary dynamic filtering mechanism in learning agents, which opens a possibility for a variety of applications in the future.

\section{References}

\bibliography{main}
\bibliographystyle{IEEEtran}

\end{document}